\pdfoutput=1
\documentclass[11pt]{article}

\usepackage{acl}

\usepackage{times}
\usepackage{latexsym}
\usepackage{hyperref}
\usepackage{fancyvrb}

\usepackage[T1]{fontenc}
\usepackage[utf8]{inputenc}

\usepackage{microtype}

\usepackage{longtable}
\usepackage{pgfplots}
\usepackage{graphicx}
\graphicspath{ {./images/} }

\title{Open-Domain Dialog Evaluation using Follow-Ups Likelihood}

\author{
    Maxime De Bruyn, Ehsan Lotfi, Jeska Buhmann, Walter Daelemans \\
    CLiPS Research Center \\
    University of Antwerp, Belgium \\
    \texttt{maxime.debruyn@uantwerpen.be}
}

\begin{document}
\maketitle
\begin{abstract}
Automatic evaluation of open-domain dialogs remains an unsolved problem. Moreover, existing methods do not correlate strongly with human annotations.
This paper presents a new automated evaluation method using follow-ups: we measure the probability that a language model will continue the conversation with a fixed set of follow-ups (e.g. \emph{Not really relevant here}, \emph{What are you trying to say?}). 
When compared against twelve existing methods, our new evaluation achieves the highest correlation with human evaluations.
\end{abstract}

\section{Introduction}

Despite the recent progress in Natural Language Processing, the automatic evaluation of open-domain conversations remains an unsolved problem. It is difficult to establish criteria to measure the quality of a system. Task-oriented dialog systems use metrics such as task success or dialog efficiency. However, these do not apply to open-domain conversational agents \cite{mctear2020conversational}.

Currently, there are two options for open-domain dialog evaluation: human evaluation and automated evaluation. Thanks to their understanding of natural language, humans are able to digest the entire dialog context in order to meaningfully evaluate a response \cite{https://doi.org/10.48550/arxiv.2203.10012}. Human evaluation also has its shortcomings: inconsistency in ratings (the same annotator may give two different scores depending on the mood), lack of reproducibility, and cost \cite{https://doi.org/10.48550/arxiv.2203.10012}. 

The second option is to use automated evaluation metrics. Methods inherited from sequence-to-sequence machine translation such as BLEU \cite{papineni-etal-2002-bleu} evaluate the generated utterance by comparing it to the ground-truth. By doing so, these methods miss the one-to-many characteristic of conversation: a conversation may evolve in more than one valid direction.

To tackle this problem, researchers came up with reference-free evaluation metrics: the generated utterance is not compared to a ground truth but evaluated on its own.

\begin{figure}[t]
    \centering
    \resizebox{\columnwidth}{!}{
    \includegraphics[width=8cm]{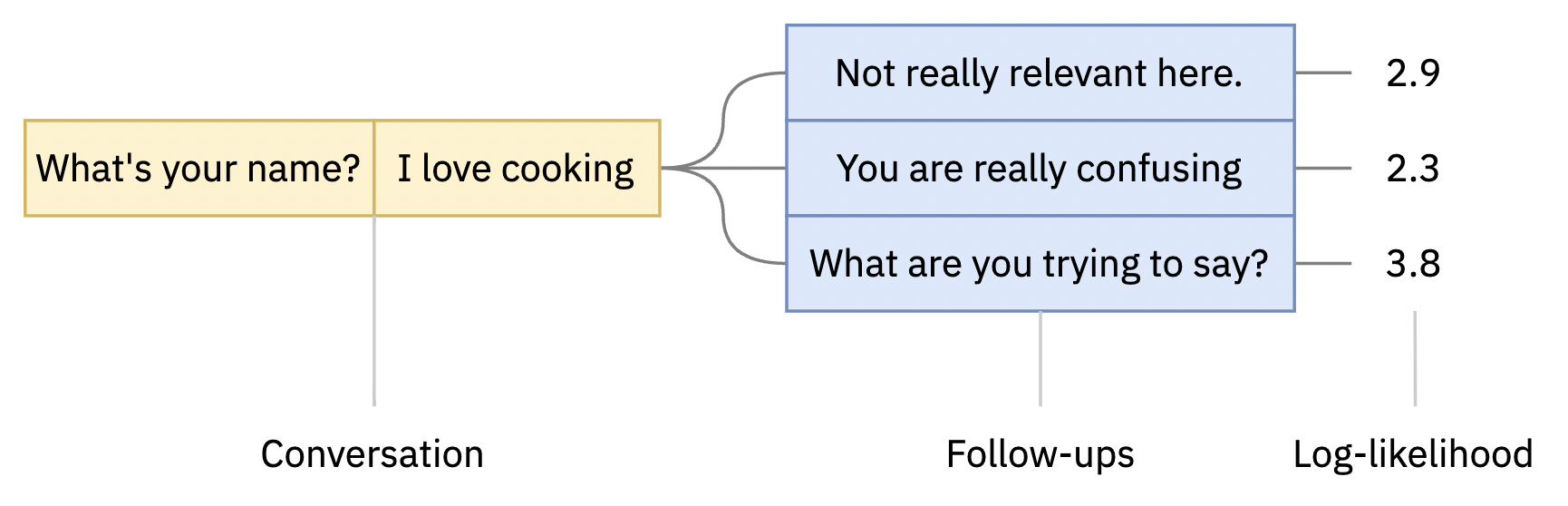}
    }
    \caption{Illustration of our method. We measure the probability (log-likelihood) that a language model will continue the conversation with a set of predefined follow-ups. This paper shows that the sum of the individual log-likelihoods correlates strongly with human evaluations.}
\end{figure}

FED \cite{mehri-eskenazi-2020-unsupervised} is an unsupervised reference-free evaluation metric. It uses the idea that one can use the next utterance in a conversation to rate the turn before it. When users speak to a system, their response to a given system may implicitly provide feedback for the system. FED uses a set of predefined follow-ups and the log-likelihood from a language model to measure 18 fine-grained attributes in a conversation.

Inspired by the FED metric, we propose a new evaluation method called FULL (Follow-Up Log-Likelihood). We start by explaining our method and how it departs from the original FED metric. Next, we explain our choice of language model and follow-ups. Finally, we demonstrate that our new method achieves the highest correlation with human evaluations compared to 12 automated metrics.
We open-source our evaluation code\footnote{\href{https://github.com/maximedb/full}{https://github.com/maximedb/full}} and publish FULL as a Python package\footnote{\href{https://pypi.org/project/full/}{https://pypi.org/project/full/}} for easy usage.

\section{Related Work}
This section reviews the existing literature on evaluation metrics for open-domain conversations. In the interest of space, we limit ourselves to studying reference-free methods (methods that do not require a ground truth). The interested reader is encouraged to read \citet{yeh-etal-2021-comprehensive} for a full review.

GRADE \cite{huang-etal-2020-grade} and DynaEval \cite{zhang-etal-2021-dynaeval} use a graph-based structure to model the dialog-level interaction between a user and a system. DynaEval distinguishes between well-formed dialogs from carefully constructed negative samples. MAUDE \cite{sinha-etal-2020-learning} is also trained to distinguish a correct response from a randomly sampled negative response using a contrastive loss. FlowScore \cite{li-etal-2021-conversations} evaluates the quality of a dialog using the dynamic information flow in the dialog history.

USR \cite{mehri-eskenazi-2020-usr} trains several models to measure different qualities of dialogs. A masked language modeling head measures the fluency of the conversation, a retrieval model determines the relevance of a response, and a fact-to-response model checks whether a response conditions on knowledge. USL-H \cite{phy-etal-2020-deconstruct} also has three internal models, although they measure different attributes: grammatical correctness, sensibleness, and the likelihood of a given response. Other notable evaluation methods include \citet{Ghazarian_Weischedel_Galstyan_Peng_2020, see-manning-2021-understanding,https://doi.org/10.48550/arxiv.2203.09711,https://doi.org/10.48550/arxiv.2203.13927}

FED \cite{mehri-eskenazi-2020-unsupervised} and HolisticEval \cite{pang-etal-2020-towards} both use GPT-like \cite{radford2019language} models to evaluate conversation on several attributes. FED computes the likelihood of manually designed follow-up utterances to measure multiple dialog qualities without supervision. HolisticEval uses a GPT-2 model to measure coherence, fluency, diversity, and consistency.

\section{Method}
\label{section:method}
Our metric FULL (Follow-Up Log-Likelihood) is a reference-free evaluation method for dialogs inspired by FED \cite{mehri-eskenazi-2020-unsupervised}. 

\subsection{Follow-Up Utterance for Evaluation}
Our method uses follow-up utterances to evaluate the quality of a conversation \cite{DBLP:journals/corr/abs-1901-06613}. When interacting with a system, users may provide implicit feedback about the conversation in the semantics of their response. For example, if a user ends a conversation with \emph{It was a pleasure talking to you}, we can reasonably assume it was a pleasant conversation. On the other hand, if a user ends a conversation with \emph{What are you talking about?}, we could conclude that the user is confused about the state of the conversation.

\subsection{Log-Likelihood of Follow-Ups}
We do not have access to the next utterance in an interactive setting. Instead, we ask a language model to play the role of a human. We ask the model how likely it is to generate a fixed set of follow-ups. For example, if the language model is likely to continue a conversation with the follow-up \emph{I don't understand what you are saying}, we could conclude that the utterance generated by the system does not make sense.

FULL analyzes the quality of a response $r$ in the context of a dialog history $h$ with a language model $M$ and a set of $n$ predefined follow-ups $F$. For each predefined follow-up, the language model computes the log-likelihood $D$ of a follow-up utterance $f_{i}$ given the dialog history. 

\begin{equation}
    \sum_{i=1}^{n}{D(h,r,f_{i})}
\end{equation}

The total score is equal to the sum of the individual log likelihoods. It is worth reminding that the metric does not mean anything. It is only useful to \emph{compare} systems together.

\subsection{Differences with FED}
Our implementation differs from FED \cite{mehri-eskenazi-2020-unsupervised} in multiple ways. First, we do not consider fine-grained attributes, only the overall quality of the turn or dialog.\footnote{Whereas FED considers 18 fine-grained attributes (overall quality included). Our initial experiments revealed that follow-ups assigned to a fine-grained attribute (e.g., engaging) often had a higher correlation with another unrelated attribute (e.g., correctness). For that reason, we choose to focus on a single attribute, the conversation's overall quality and leave the study of fine-grained attributes for future work.}

Second, FED computes the log-likelihood of the conversation history $h$, the response $r$, and the follow-up $f_{i}$. Whereas we only compute the conditional log-likelihood of the follow-up $f_{i}$. Computing the log-likelihood over the conversation introduces a bias towards the dataset used in training the language model, Reddit, in the case of FED. It also favors longer conversations over shorter ones. Our goal is to estimate the likelihood of the follow-up, not the conversation itself.

Third, FED did not justify its choice of follow-ups, while we studied each candidate and only took the most correlated ones making intuitive sense. Fourth, we also study multiple types of language models (conversational and general).

\section{Experimental Settings}
\label{section:experimentalsettings}
This section explains our choices of follow-ups, language models, and conversational data. Our goal is to find the combination of language models and follow-ups correlating the most with human evaluations.

\subsection{Follow-Ups}
A follow-up is an utterance added after a conversation's last turn to evaluate the last turn or the entire dialog. FED defined 63 unique follow-ups in 16 categories (fine-grained attributes) at the turn level and the dialog level. Appendix \ref{appendix:list_followups} list the entire list of follow-ups. The authors did not provide any justification for their choice of follow-ups. Instead of blindly using the list of follow-ups, we attempt to understand which of these follow-ups have the highest correlation with human evaluations.

\subsection{Language Models}
We experiment with several language models, both general and conversational. The goal of the language module is to compute the conditional log-likelihood of several follow-ups.
\paragraph{BlenderBot v1} is a conversational sequence-to-sequence model \cite{DBLP:journals/corr/abs-2004-13637} with three sizes: small, large, and extra-large. A distilled version is also available on HuggingFace.
\paragraph{DialoGPT} is a conversational language model \cite{zhang-etal-2020-dialogpt} with three sizes: small, medium and large. The authors fine-tuned a GPT-2 model on a large corpus of Reddit conversations.
\paragraph{GPT-2} is a general language model \cite{radford2019language}. While it was not trained specifically on conversational data, our experiments revealed its potential to estimate a conversation's quality.

\subsection{Conversational Data}
We use the FED dataset \cite{mehri-eskenazi-2020-unsupervised} for evaluating the set of follow-ups. It consists of 372 turn-level (124 dialog-level), originally collected by \citet{DBLP:journals/corr/abs-2001-09977}. The dataset consists of human-system conversations (Meena and Mitsuku) and human-human conversations. \citet{mehri-eskenazi-2020-unsupervised} asked annotators to evaluate turn-level and dialog-level conversations on several attributes. In this work, we only use the evaluation of the overall quality of the turn or dialog.

\section{Results}
\begin{table}
\centering
\resizebox{\columnwidth}{!}{
\begin{tabular}{lcc}
\hline
\textbf{Follow-up}                     & \multicolumn{2}{c}{\textbf{Correlation}} \\ \hline
                                       & \textbf{Turn}      & \textbf{Dialog}     \\ \hline
Not really relevant here.              & 0.48               & 0.65                \\
You're really confusing.               & 0.46               & 0.67                \\
I don't understand what you're saying. & 0.46               & 0.58                \\
That's not really relevant here.       & 0.45               & 0.70                \\
You are so confusing.                  & 0.45               & 0.64                \\
You're really boring.                  & 0.44               & 0.65                \\
That's not very interesting.           & 0.44               & 0.60                \\
That was a really boring response.     & 0.43               & 0.63                \\
You don't seem interested.             & 0.43               & 0.61                \\
I am so confused right now.            & 0.43               & 0.57                \\ \hline
\end{tabular}
}
\caption{Top 10 follow-ups ranked by Spearman correlation to human evaluations. All follow-ups exhibit a positive relationship, meaning that the likely presence of the follow-up (low log-likelihood) entails a low human evaluation and vice-versa.}
\label{table:correlation}
\end{table}
Our objective is to find the best combination of language models and follow-ups. We start by analyzing which language model correlates the most with human evaluation. In the second step, we look for the best set of follow-ups.

\subsection{Choice of Language Model}
We are looking for a language model whose log-likelihood of generating the follow-ups correlates highly with human evaluations. We do so both on a turn-level and dialog-level. We compare the average absolute correlation of each follow-up with human judgments. The results are displayed on Figure \ref{figure:avgcorrelation} in Annex \ref{appendix:model_comparison}. The model standing out is the large Blender model \cite{DBLP:journals/corr/abs-2004-13637}. It has the highest correlation with humans both on a turn-level and dialog-level. The difference in performance between Blender-3B and Blender-400M is small. For these reasons, we choose Blender-400M as our default language model.

\subsection{Choice of Follow-ups}
Now that we have identified our model of choice (Blender-400M), we wish to identify the follow-ups correlating the most with humans. We compute the Spearman correlation between each follow-up and human evaluation (turn-level and dialog-level). We present the top-10 follow-ups (by absolute correlation) in Table \ref{table:correlation}. The full table is available 
 Appendix \ref{appendix:list_followups}.

The follow-up correlating the most on a turn-level basis is \emph{Not really relevant here} with a Spearman correlation of 0.48. The least correlated follow-up is \emph{Wow! That's really cool!} with correlations of 0.04. The follow-up correlating the most on a dialog-level basis is \emph{That's not really relevant here} with a correlation of 0.70. The least correlated follow-up on a dialog level is \emph{Cool! That sounds super interesting!} with a correlation of 0.01. 

Most follow-ups exhibit a positive relationship, meaning that the likely generation of the follow-up by the language model (low log-likelihood) entails a low human rating and vice-versa. However, all the top follow-ups are \emph{negative} follow-ups (e.g., \emph{You're really confusing}), and their likely presence indicates a negative conversation. On the other hand, the \emph{positive} follow-ups (e.g., \emph{Great talking to you}) are not as highly correlated.
On average, negative follow-ups correlate with 0.39, while positive follow-ups correlate with 0.24. These results indicate that the language model evaluates a good conversation by the likely absence of negative follow-ups.

\begin{table}
\resizebox{\columnwidth}{!}{
\begin{tabular}{lcc}
\hline
                 & \textbf{Turn Level} & \textbf{Dialog Level} \\ \hline
QuestEval        & 0.09                & 0.08                  \\
MAUDE            & -0.09               & -0.28                 \\
DEB              & 0.19                & -0.01                 \\
GRADE            & 0.12                & -0.06                 \\
DynaEval         & 0.32                & 0.55                  \\
USR              & 0.12                & 0.06                  \\
USL-H            & 0.19                & 0.15                  \\
DialoRPT         & -0.09               & -0.21                 \\
HolisticEval     & 0.12                & -0.30                 \\
PredictiveEngage & 0.09                & 0.15                  \\
FED              & 0.09                & 0.32                   \\
FlowScore        & -0.05               & -0.00                 \\ \hline
FULL (ours)             & \textbf{0.51}       & \textbf{0.69}         \\ \hline
\end{tabular}
}
\caption{Comparison of our evaluation method FULL with other automated methods. FULL achieves the highest correlation on turn-level and dialog-level, followed by DynaEval. Except for FULL, results are copied from \citet{yeh-etal-2021-comprehensive}.}
\label{table:comparison}
\end{table}

Each follow-up brings another forward pass of the model, so ideally, we want to restrict the number of follow-ups in the final evaluation method. For the final selection of follow-ups, we combine the rank of the turn-level and dialog-level correlations and take the top 5.\footnote{We arbitrarily choose the number 5. We also removed close duplicates. For example \emph{Not really relevant here.} and \emph{That's not really relevant here.}} The final selection of follow-ups is the following: \emph{Not really relevant here.} \emph{You're really confusing.} \emph{You're really boring.} \emph{What are you trying to say?} \emph{You don't seem interested.}

\subsection{Comparison}
 \citet{yeh-etal-2021-comprehensive} compared 12 evaluation methods on the FED dataset \cite{mehri-eskenazi-2020-unsupervised}. We compare our method FULL against these 12 other methods in Table \ref{table:comparison}. The results are clear, FULL achieves the highest correlation both on a turn-level and dialog-level while being fully unsupervised (except in the choice of follow-ups). By combining the log-likelihood from 5 follow-ups, the average correlation on turn-level increases to 0.51, while the average of the individual correlation equals 0.45. 

\section{Conclusion}
This short paper introduces a new automated evaluation method (FULL) for open-domain conversations. FULL measures the quality of a conversation by computing the probability that a language model will continue the conversation with a set of follow-ups (e.g., \emph{Not really relevant here}, \emph{What are you trying to say?}). FULL achieves the highest correlation with human evaluations compared to twelve other existing methods.

Our experiments revealed that negative follow-ups (e.g., \emph{Not really relevant here}) have a higher correlation with human evaluations than positive follow-ups (e.g., \emph{Wow, interesting to know}). It is easier for the model to evaluate a conversation from its bad angles rather than its good ones.

Future work is needed to know which fine-grained attribute can be measured using the same technique. Using ever-large models such as GPT-3 \cite{DBLP:journals/corr/abs-2005-14165} or OPT \cite{https://doi.org/10.48550/arxiv.2205.01068} could be a direction for future research, although the resulting model will likely need to be distilled to be of practical use.

\bibliography{anthology,custom}
\bibliographystyle{acl_natbib}

\appendix
\section{Appendix: Comparison of Models}
\label{appendix:model_comparison}
\begin{figure*}
\centering
\resizebox{\textwidth}{!}{
\begin{tikzpicture}
\begin{axis}[
    ybar,
    enlargelimits=0.1,
    width=16cm,
    height=5cm,
    symbolic x coords={{BLD S},{BLD L},{DGPT S},{DGPT M},{DGPT L},{GPT2 S},{GPT2 M},{GPT2 L},{GPT2 XL}},
    bar width=6pt,
    %nodes near coords,
    xtick=data,
    legend style={at={(0.5,-0.2)},
		anchor=north,legend columns=-1},
    label style={font=\small},
    tick label style={font=\small} 
]
    \addplot 
        coordinates {({BLD S},33.6) ({BLD L},33.4) ({DGPT S},16.4) ({DGPT M},15.9) ({DGPT L},15.3) ({GPT2 S},17.2) ({GPT2 M},20.3) ({GPT2 L},20.0) ({GPT2 XL},20.0)};
    \addplot 
	    coordinates {({BLD S},40.6) ({BLD L},43.1) ({DGPT S},24.6) ({DGPT M},19.8) ({DGPT L},19.5) ({GPT2 S},25.9) ({GPT2 M},32.5) ({GPT2 L},30.5) ({GPT2 XL},33.4)};
\legend{Per Turn,Entire Dialog}
\end{axis}
\end{tikzpicture}
}
\caption{Average absolute correlation with human evaluations for several language models. We use  Blender-400 (BLD S) as language model because of its high correlation with human evaluations. For space reasons, Blender is abbreviated as BLD and DialoGPT as DGPT.}
\label{figure:avgcorrelation}
\end{figure*}
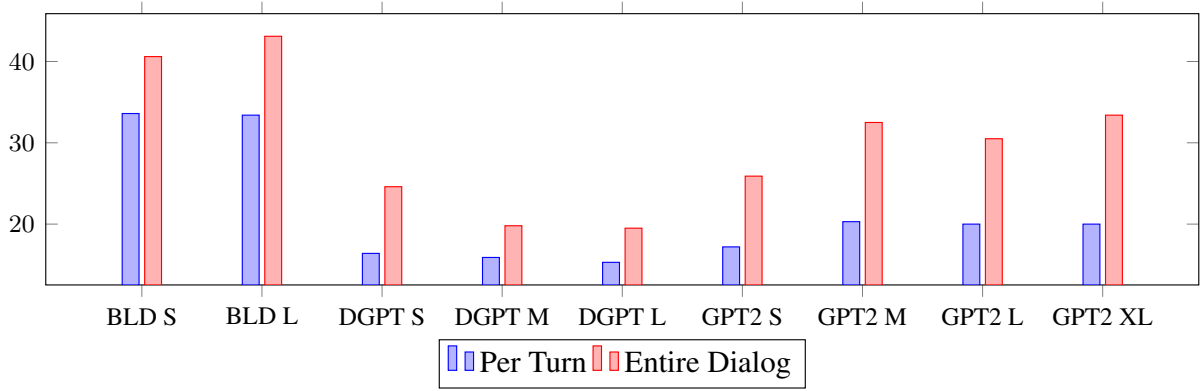

We present in Figure \ref{figure:avgcorrelation} the average absolute correlation to human evaluations per model.

\section{Appendix: List of Candidate Follow-ups}
\label{appendix:list_followups}
Table \ref{tab:list-of-followups} list the entire list of follow-ups considered.

\onecolumn

\begin{longtable}{lp{5.5cm}lllrl}
\hline
 & \textbf{Follow-up} & \textbf{Category} & \textbf{Level} & \textbf{Type} & \multicolumn{1}{c}{\textbf{Level}} & \multicolumn{1}{c}{\textbf{Dialog}} \\ \hline
\endfirsthead
\multicolumn{7}{c}%
{{\bfseries Table \thetable\ continued from previous page}} \\
\hline
 & \textbf{Follow-up} & \textbf{Category} & \textbf{Level} & \textbf{Type} & \multicolumn{1}{c}{\textbf{Level}} & \multicolumn{1}{c}{\textbf{Dialog}} \\ \hline
\endhead
\cline{2-7}
\endfoot
\endlastfoot
X & Not really relevant here. & specific & turn & neg & 0.48 & 0.65 \\
X & You're really confusing. & error recovery & dialog & neg & 0.46 & 0.67 \\
 & I don't understand what you're saying. & correct & turn & neg & 0.46 & 0.58 \\
 & That's not really relevant here. & specific & turn & neg & 0.45 & 0.70 \\
 & You are so confusing. & coherent & dialog & neg & 0.45 & 0.64 \\
X & You're really boring. & informative & dialog & neg & 0.44 & 0.65 \\
 & That's not very interesting. & interesting & turn & neg & 0.44 & 0.60 \\
 & That was a really boring response. & interesting & turn & neg & 0.43 & 0.63 \\
X & You don't seem interested. & inquisitive & dialog & neg & 0.43 & 0.61 \\
 & I am so confused right now. & error recovery & dialog & neg & 0.43 & 0.60 \\
 & I'm so confused! & understandable & turn & neg & 0.43 & 0.59 \\
 & I don't really care. That's pretty   boring. & engaging & turn & neg & 0.43 & 0.61 \\
 & I want to talk about something else. & engaging & turn & neg & 0.43 & 0.65 \\
 & That's not even related to what I said. & relevant & turn & neg & 0.42 & 0.58 \\
X & What are you trying to say? & understanding & dialog & neg & 0.42 & 0.68 \\
 & I am so confused right now! & correct & turn & neg & 0.42 & 0.57 \\
 & That makes no sense! & semantically appropriate & turn & neg & 0.42 & 0.56 \\
 & I don't understand at all! & understandable & turn & neg & 0.41 & 0.54 \\
 & That's really boring. & interesting & turn & neg & 0.41 & 0.54 \\
 & I don't like you. & likeable & dialog & neg & 0.40 & 0.58 \\
 & I'm so confused right now! & fluent & turn & neg & 0.40 & 0.56 \\
 & Don't change the topic! & relevant & turn & neg & 0.40 & 0.58 \\
 & You're not understanding me! & correct & turn & neg & 0.40 & 0.62 \\
 & That's a very generic response. & specific & turn & neg & 0.39 & 0.50 \\
 & You don't really know much. & informative & dialog & neg & 0.39 & 0.52 \\
 & You're not very nice. & likeable & dialog & neg & 0.38 & 0.56 \\
 & You're not very fun to talk to. & likeable & dialog & neg & 0.37 & 0.55 \\
 & Is that real English? & fluent & turn & neg & 0.37 & 0.49 \\
 & That's a lot of questions! & inquisitive & dialog & pos & 0.36 & 0.52 \\
 & Why are you repeating yourself? & diverse & dialog & neg & 0.35 & 0.50 \\
 & You're making no sense at all. & coherent & dialog & neg & 0.35 & 0.43 \\
 & You ask a lot of questions! & inquisitive & dialog & pos & 0.35 & 0.54 \\
 & Let's change the topic. & engaging & turn & neg & 0.35 & 0.45 \\
 & You don't ask many questions. & inquisitive & dialog & neg & 0.35 & 0.54 \\
 & Why are you changing the topic? & relevant & turn & neg & 0.34 & 0.51 \\
 & Stop saying the same thing repeatedly. & diverse & dialog & neg & 0.34 & 0.50 \\
 & Do you know how to talk about something   else? & flexible & dialog & neg & 0.33 & 0.49 \\
 & You're changing the topic so much! & coherent & dialog & neg & 0.33 & 0.47 \\
 & You know a lot of facts! & informative & dialog & pos & 0.32 & 0.48 \\
 & Tell me more! & engaging & turn & pos & 0.32 & 0.34 \\
 & I like you! & likeable & dialog & pos & 0.31 & 0.43 \\
 & Wow that's a lot of information. & informative & dialog & pos & 0.31 & 0.38 \\
 & Stop changing the topic so much. & depth & dialog & neg & 0.31 & 0.44 \\
 & What does that even mean? & understandable & turn & neg & 0.30 & 0.35 \\
 & I don't want to talk about that! & flexible & dialog & neg & 0.29 & 0.50 \\
 & That's not what you said earlier! & consistent & dialog & neg & 0.29 & 0.37 \\
 & You have a good point. & semantically appropriate & turn & pos & 0.29 & 0.43 \\
 & I see, that's interesting. & specific & turn & pos & 0.28 & 0.31 \\
 & Stop contradicting yourself! & consistent & dialog & neg & 0.28 & 0.36 \\
 & You're very easy to talk to! & flexible & dialog & pos & 0.28 & 0.40 \\
 & Stop repeating yourself! & diverse & dialog & neg & 0.27 & 0.40 \\
 & That's good to know. Cool! & specific & turn & pos & 0.25 & 0.30 \\
 & That's a good point. & specific & turn & pos & 0.25 & 0.34 \\
 & Wow you can talk about a lot of things! & flexible & dialog & pos & 0.23 & 0.27 \\
 & I'm really interested in learning more   about this. & engaging & turn & pos & 0.22 & 0.26 \\
 & That makes sense! & semantically appropriate & turn & pos & 0.21 & 0.21 \\
 & Thanks for all the information! & informative & dialog & pos & 0.21 & 0.15 \\
 & You're super polite and fun to talk to & likeable & dialog & pos & 0.17 & 0.23 \\
 & Wow that is really interesting. & interesting & turn & pos & 0.17 & 0.14 \\
 & That's really interesting! & interesting & turn & pos & 0.16 & 0.11 \\
 & Great talking to you. & likeable & dialog & pos & 0.15 & 0.10 \\
 & Cool! That sounds super interesting. & interesting & turn & pos & 0.08 & -                       0.01 \\
 & Wow! That's really cool! & engaging & turn & pos & 0.04 & -                       0.08 \\ \cline{2-7} 
\caption{List of candidate follow-ups along with their category (fine-grained attribute), positivity (negative of positive follow-up) and correlation with a human evaluation of the overall quality of the turn/dialog. All follow-ups and static data is from \citet{mehri-eskenazi-2020-unsupervised}.}
\label{tab:list-of-followups}
\end{longtable}
\twocolumn

\clearpage

\end{document}